\documentclass{vgtc}                          

\graphicspath{{figures/}{pictures/}{images/}{./}} 

\usepackage{times}                     

\usepackage{tabu}                      
\usepackage{booktabs}                  
\usepackage{lipsum}                    
\usepackage{mwe}                       

\usepackage{amsmath, amssymb}
\usepackage[capitalize]{cleveref}

\usepackage{amsmath}
\usepackage{multirow}

\newcommand{\ie}{\textit{i}.\textit{e}.}
\newcommand{\eg}{\textit{e}.\textit{g}.}

\usepackage{graphicx}
\usepackage{color}
\definecolor{mycolor1}{RGB}{255,153,153}
\definecolor{mycolor2}{RGB}{255,204,153}
\usepackage{colortbl}  
\usepackage{xcolor}
\usepackage{array}   
\usepackage{makecell}
\usepackage{bm}
\newcommand{\vm}[1]{\mathbf{#1}}
\newcommand{\ve}[1]{\mbox{\bm{$#1$}}}

\usepackage{pifont}

\usepackage{mathptmx}                  

\onlineid{1299}

\vgtccategory{Research}

\title{ExFMan: Rendering 3D Dynamic Humans with Hybrid Monocular \\Blurry Frames and Events}

\author{Kanghao Chen$^{1}$ \quad Zeyu Wang$^{2,1}$ \quad Lin Wang$^{1,2,3}$\thanks{Corresponding Author}\\
$^{1}$AI Thrust, $^{2}$CMA Thrust, HKUST(GZ) $^{3}$Dept. of CSE, HKUST \\
{kchen879@connect.hkust-gz.edu.cn, linwang@ust.hk}}

\teaser{
  \centering
  \includegraphics[width=\linewidth]{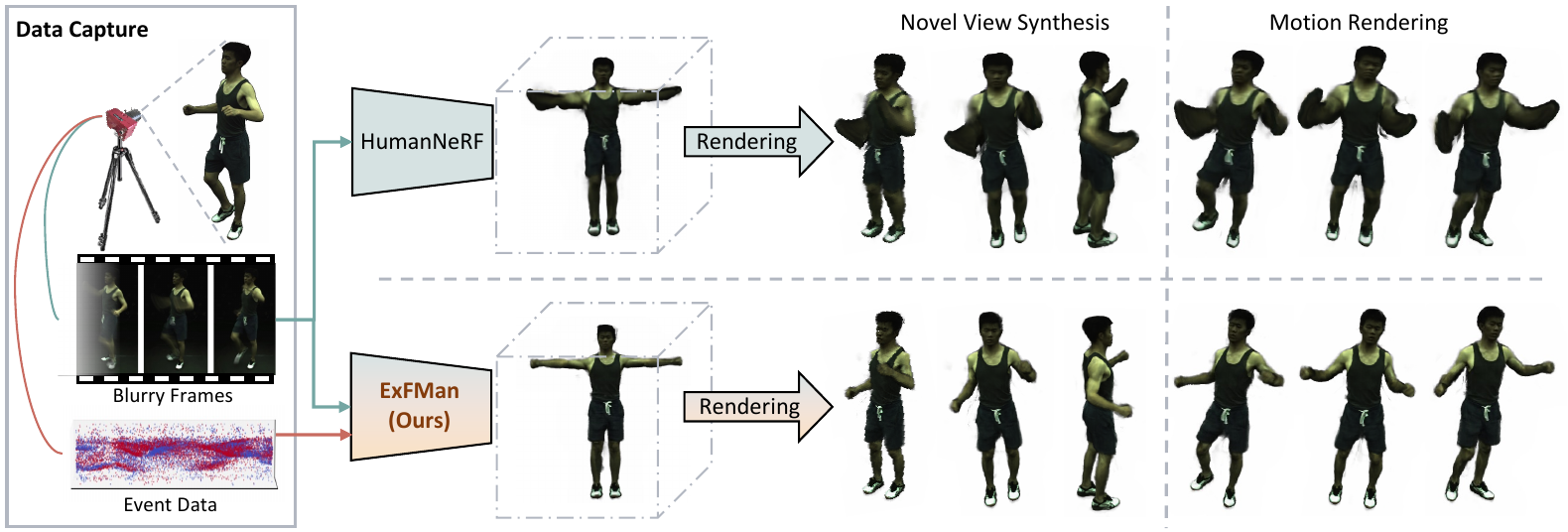}
  \vspace{-20pt}
  \caption{Dynamic human motion often causes severe blur in the videos, imposing additional challenges into the optimization process. \textbf{Left}: Data capture system to obtain simultaneously recording spatial-temporally aligned event data and RGB frames with humans performing complex and rapid motion.  \textbf{Top row}: baseline method optimizes only on the RGB without considering the motion blur, which results in blurry rendered results; \textbf{Bottom row}: Our method incorporates RGB and event data to optimize the human in a novel pipeline and render high-quality results.}
  \vspace{-10pt}
  \label{fig:overview}
}

\abstract{
Recent years have witnessed tremendous progress in the 3D reconstruction of dynamic humans from a monocular video with the advent of neural rendering techniques.
This task has a wide range of applications, including the creation of virtual characters for virtual reality (VR) environments.
However, it is still challenging to reconstruct clear humans when the monocular video is affected by motion blur, particularly caused by rapid human motion (\eg, running, dancing), as often occurs in the wild.
This leads to distinct inconsistency of shape and appearance for the rendered 3D humans, especially in the blurry regions with rapid motion, \eg, hands and legs.
In this paper, we propose \textbf{ExFMan}, the \textit{first} neural rendering framework that unveils the possibility of rendering high-quality humans in rapid motion with a hybrid frame-based RGB and bio-inspired event camera.
The ``out-of-the-box'' insight is to leverage the high temporal information of event data in a complementary manner and adaptively reweight the effect of losses for both RGB frames and events in the local regions, according to the velocity of the rendered human.
This significantly mitigates the \textit{inconsistency} associated with motion blur in the RGB frames.
Specifically, we first formulate a velocity field of the 3D body in the canonical space and render it to image space to identify the body parts with motion blur. We then propose two novel losses, \ie, velocity-aware photometric loss and velocity-relative event loss, to optimize the neural human for both modalities under the guidance of the estimated velocity. In addition, we incorporate novel pose regularization and alpha losses to facilitate continuous pose and clear boundary.
Extensive experiments on synthetic and real-world datasets demonstrate that ExFMan can reconstruct sharper and higher quality humans over the compared baselines and the state-of-the-art methods for diverse blurry subjects.

} 


\begin{document}


\firstsection{Introduction}

\maketitle

\label{sec:intro}
The digital modeling of humans holds immense potential for the development of innovative applications in extended reality (XR)~\cite{rendle2023volumetric,morgenstern2024animatable,ren2023lidar}, where human models can serve as natural and intuitive interfaces. Achieving dynamic modeling and rendering of 3D humans is essential, as individuals are highly sensitive to the intricate appearance of human bodies and faces.
In particular, rendering humans from in-the-wild monocular videos has triggered significant interest due to the widespread availability of videos from a monocular camera and the broad demand in applications for complex scenes~\cite{weng2022humannerf}.
Recently, some pioneering works~\cite{peng2021neural,peng2021animatable,hu2024gauhuman} are proposed to learn dynamic humans based on the body poses and observation frames by deforming the neural radiance fields (NeRF)~\cite{mildenhall2021nerf} and 3D gausian splatting (3DGS)~\cite{kerbl20233d}. 
Some follow-up research efforts are then made to enhance the rendering quality~\cite{peng2021neural, weng2022humannerf,jiang2022neuman} and optimize the rendering efficiency~\cite{peng2022selfnerf, jiang2023instantavatar, kocabas2023hugs,lei2023gart}. 
While these methods have demonstrated the capability to render humans from novel views accurately, they are primarily tailored to controlled environments with well-designed actions of humans and ideal illumination conditions.
In the in-the-wild scenarios, particularly when subjects engage in free-form actions under varying lighting conditions, these methods often encounter motion blur across video frames~\cite{zamir2021multi}. Motion blur leads to a deterioration in the quality of affected regions, causing a loss of fine details and introducing inconsistencies in these areas during the optimization of the human body, as demonstrated in \cref{fig:overview}(top).

Therefore, it is imperative to reconstruct 3D dynamic humans from the in-the-wild monocular blurry video; however, this task is non-trivial for a key reason -- \textbf{inconsistency of human shape and appearance}. Some methods~\cite{weng2022humannerf,yu2023monohuman} optimize the dynamic human in a single static canonical space and deform it to fit multiple frames, which essentially relies on the underlying assumption of consistency of overall frames~\cite{mildenhall2021nerf}.  Although some methods~\cite{ma2022deblur, lee2023exblurf, wang2023bad, qi2023e2nerf} reconstruct NeRF from blurry frames, they mainly focus on the general scene and only consider the blur from camera shift or defocus. This renders it difficult to apply them to reconstruct dynamic humans directly. To our knowledge, \textit{\textbf{no prior work has directly addressed the challenge of rendering dynamic humans from the monocular blurry video}}.
A naive solution is to combine the video deblurring methods, \eg~, MPR~\cite{zamir2021multi} with the human rendering methods, \eg~, HumanNeRF~\cite{weng2022humannerf} based on the enhanced frames in a two-stage manner. However, such a solution heavily relies on deblurring methods, which suffer from limited generalization capability across different scenes and often fail with complex motion. Consequently, it leads to error-prone appearance and shape in the rendered humans, as demonstrated in \cref{fig:zju_mocap_main}.


Event cameras are bio-inspired sensors that capture per-pixel intensity changes asynchronously. Recently, they are becoming more and more popular for their distinct advantages, \eg, high temporal resolution and high dynamic range (HDR). 
Accordingly, event cameras have been applied to tasks like video deblurring~\cite{jiang2020learning, lin2020learning, pan2019bringing, shang2021bringing, xu2021motion} and low-light enhancement~\cite{chen2024evlight++,liang2023coherent}. Moreover, they have been leveraged to capture 3D human motion~\cite{zou2021eventhpe, xu2020eventcap} or estimate the human pose and shape~\cite{jiang2023evhandpose, rudnev2021eventhands, nehvi2021differentiable}, by leveraging the advantages of event data.


In this paper, we explore a promising direction of rendering 3D dynamic humans with hybrid monocular blurry RGB frames and events. 
Intuitively, we introduce \textbf{ExFMan}, the \textbf{\textit{first}} neural rendering framework that can render dynamic humans with high-quality appearance under diverse blur conditions. Our method 
can achieve high-quality novel view synthesis and motion rendering, as depicted in \cref{fig:overview}.
The key insight is to leverage the high temporal information of events in a complementary manner and adaptively reweight the effect of losses for both RGB frames and events in the local regions, according to the velocity of the rendered human.
This significantly mitigates the \textit{\textbf{inconsistency}} associated with motion blur in the RGB frames. 
To achieve this, we first formulate a velocity field of a 3D body within the canonical space, which is then deformed and rendered to the image space to identify the body regions with motion blur (\cref{sec:velocity_field}).  

Based on the estimated velocity, we propose two velocity-based rendering losses to facilitate the joint optimization for both modalities (\cref{sec:rendering_loss}).  Specifically, a velocity-aware photometric loss is designed by representing the rendered color through a Gaussian distribution, with the velocity score representing the variance. This approach directly diminishes the impact of the blurry regions, thus preserving the consistency of the human body across video frames.
The body regions (\eg, hands, legs) that always exhibit motion blur are not adequately optimized by the proposed photometric loss. Therefore, we apply additional supervision by designing a velocity-relative event loss for these under-constrained regions. 
Regarding the velocity as prediction uncertainty, the novel event loss optimizes the high-velocity regions by anchoring them to regions with lower velocity, with the actual event data as supervision.
Additionally, we incorporate the pose regularization and velocity-based alpha loss to facilitate the continuous pose and clear boundary (\cref{sec:optimization}).

In summary, our contributions are summarized as follows:
\begin{itemize}
    \item We present the \textbf{first} work for rendering dynamic humans in rapid motion with hybrid monocular blurry RGB frames and sparse events, leveraging the unique capabilities of event cameras. It significantly enhances the robustness of rendering dynamic humans in uncontrolled, real-world environments.
    \item We present a novel framework that involves an event-oriented blur-aware velocity field and two velocity-aware rendering losses to enable human rendering from blurry videos.
    \item We show that our method significantly outperforms the compared baselines and the state-of-the-art methods in complex blurry scenes. ExFMan achieves a \textbf{3.06 dB} PSNR improvement compared with the comparison methods on the ZJU-Mocap~\cite{peng2021neural} dataset. It shows a promising direction that could inspire more follow-up research.
\end{itemize}

\begin{figure*}[t]
    \centering
    
    \includegraphics[width=1\linewidth]{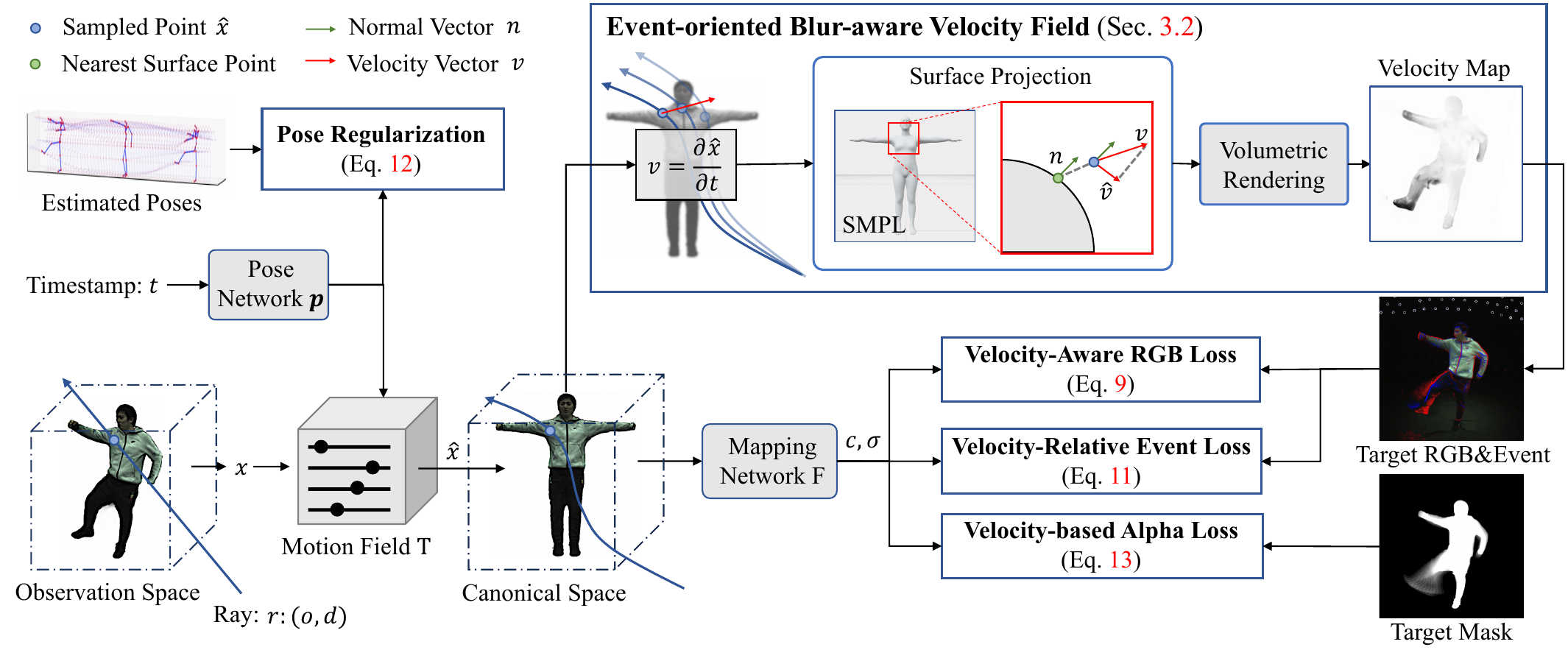}
    \vspace{-20pt}
    \caption{\textbf{Our framework facilitates human reconstruction from blurry frames and event data, based on a velocity field.} For a given timestamp $t$ and 3D point $\ve{x}$ along the sampled ray $\ve{r}$ in the observation space, the deformation mapping deforms $\ve{x}$ into its canonical counterpart $\hat{\ve{x}}$, according to pose $\ve{p}(t)$. The velocity $\ve{v}$ of sampled point $\ve{x}$ and time $t$ is calculated as the derivative of the deformed point $\hat{\ve{x}}$ w.r.t the timestamp $t$, followed by projecting it onto the human surface based on normal vector $\ve{n}$ of SMPL model. The velocity map is then obtained through volumetric rendering. Based on this velocity map, we introduce a velocity-aware photometric loss and a velocity-relative event loss, designed to leverage both data modalities in a complementary manner.}
    \vspace{-6pt}
    \label{fig:framwork}
\end{figure*}

\section{Related Work}
\subsection{Neural Rendering for Human Reconstruction.}
Following the emergence of Neural Radiance Fields (NeRF)~\cite{mildenhall2021nerf}, a variety of advancements have been made to facilitate the high-fidelity rendering of static scenes~\cite{barron2022mip, verbin2022ref, tancik2020fourier, sun2022direct, muller2022instant}, moving subjects~\cite{gao2021dynamic,li2021neural,park2021nerfies,park2021hypernerf,pumarola2021d,ost2021neural}, and dynamic humans~\cite{feng2022capturing,jiang2022selfrecon,li2022tava,liu2021neural,peng2021animatable,wang2022arah,xu2021h,jiang2023instantavatar,jiang2022neuman}. NeRF predicts the color and density for each sampled point along a ray in 3D space, integrating them through volume rendering. This technique allows for the reproduction of complex lighting and texture details, which are often challenging to replicate using conventional methods. Recent works have facilitated NeRF to human reconstruction for various applications and scenarios, such as monocular video~\cite{weng2022humannerf, jiang2022neuman}, rendering efficiency~\cite{lin2022efficient,chen2023fast,jiang2023instantavatar}, animation~\cite{yu2023monohuman}, occluded modelling~\cite{xiang2023rendering,xiang2023wild2avatar}. Our research builds upon HumanNeRF~\cite{weng2022humannerf}, recognized for its exceptional rendering capabilities for monocular video. HumanNeRF employs a static T-pose human as the canonical model and develops a deformation mapping~\cite{weng2020vid2actor} to transition this canonical form to the corresponding observation space for each video frame (details are provided in \cref{sec:preliminary}).
Recently, there has been a surge of interest in the 3D Gaussian splitting (3DGS) representation~\cite{kerbl20233d} due to its ability to achieve a balance between real-time rendering speed and photorealistic rendering quality. The field of 3D Gaussian-based avatar reconstruction~\cite{hu2024gauhuman,jung2023deformable,li2023human101,kocabas2023hugs,lei2023gart} has experienced rapid growth and become a bustling
area of research within a short period of time.
However, previous works typically rely on clean training data, assuming well-designed human motion in monocular videos without motion blur. When applied to in-the-wild scenarios, these methods struggle to recover the clear structure of humans. It also doesn't matter what kind of representation is used (See \cref{fig:zju_mocap_main}).
Differently, \textit{our work models the velocity field to effectively localize the blur degradation and render a dynamic human from monocular blurry frames and events based on novel rendering losses.}

\subsection{Deblurring NeRF/3DGS.}
Various methods~\cite{wu2022dof, ma2022deblur,lee2023dp,peng2022pdrf,chen2024deblur} have been developed to adapt NeRF/3DGS for generating clear and sharp outputs from blurry inputs. Deblur-NeRF~\cite{ma2022deblur} pioneers deblurring within NeRF without requiring clear images during its training phase. It utilizes an ancillary, compact MLP to estimate a per-pixel blur kernel. 
Subsequent advancements, such as DP-NeRF~\cite{lee2023dp}, have enhanced image quality, and PDRF~\cite{peng2022pdrf} has introduced grid-based modifications to the MLP, improving rendering speed. 
Deblur-GS~\cite{chen2024deblur} formulates a camera motion deblurring method for 3D Gaussian scene representation.
Nonetheless, \textit{the existing method mainly focuses on static scenes and considers the blur caused by camera shift or defocus, which can not be directly applied to the blur caused by the motion of the human body}.

Thanks to the robustness of event cameras toward motion blur, recent works~\cite{hwang2023ev,rudnev2023eventnerf,xiong2024event3dgs} have been proposed to optimize NeRF/3DGS derived only from the event stream. Recent works~\cite{qi2023e2nerf,cannici2024mitigating} also combine events and images for NeRF-based reconstruction to alleviate the blur caused by extreme camera shake.
Moreover, DE-NeRF~\cite{ma2023deformable} models dynamic scenes using the event and RGB camera. It focuses on capturing the fast-deforming radiance field based on sharp frames and events, which is different from our motivation of dynamic human reconstruction from monocular blurry frames.
\textit{We are the first to explore the potential of event cameras and use them as guidance for tackling the challenges of rendering dynamic humans in diverse blurry scenes.} In addition, while the previous methods roughly combine the RGB frames and event data in the optimization pipeline, our method proposes to leverage both modalities in a complementary way by considering the unique characteristics of human modeling.

\subsection{Video Deblurring.}
Deblurring remains a complex challenge in image restoration, primarily due to its inherent ill-posed nature. Traditional approaches~\cite{deconvolution2007blind,cho2009fast,gupta2010single,hyun2014segmentation} involve estimating a 2D blur kernel results in the blurry image. With the advent of deep learning and the availability of extensive datasets~\cite{nah2019ntire,nah2017deep,rim2022realistic,rim2020real,zhong2020efficient} containing pairs of blurry and sharp images, recent strategies~\cite{chakrabarti2016neural, cho2021rethinking,tao2018scale,wieschollek2017learning,zhang2020deblurring} have relied supervised learning with convolutional neural networks (CNN). Events have been involved for motion deblurring in recent research~\cite{pan2019bringing,shang2021bringing,
sun2022event}, due to the strong connection they possess with motion information. For the task of human reconstruction, these methods can serve as a \textit{two-stage baseline}, first deblurring and then reconstructing the 3D human. However, most of the methods not only suffer from limited generalization capability across different scenes but also ignore the human-specific prior in the problem, which often fails with complex motion. Therefore, directly applying them to our problem leads to error-prone appearance and shape in the rendered humans (See \cref{fig:zju_mocap_main}). 


\section{Methodology}
In this section, we describe our ExFMan that can reconstruct dynamic humans from hybrid blurry frames and events.
An overview of our framework is shown in \cref{fig:framwork}. 
We first review preliminaries and background about HumanNeRF~\cite{weng2022humannerf} and event supervision used for optimization in \cref{sec:preliminary}. We then delineate the formulation of the event-oriented blur-aware velocity field of a 3D human in \cref{sec:velocity_field}. Based on the estimated velocity, we then describe the proposed two novel rendering losses, namely the velocity-aware photometric loss and velocity-relative event loss in \cref{sec:event_loss}. Finally, pose regularization and velocity-based blur alpha loss are introduced and jointly optimized with other loss terms in \cref{sec:optimization}.

\subsection{Preliminaries and Background}

\begin{figure*}[t]
    \centering
    \includegraphics[width=1\linewidth]{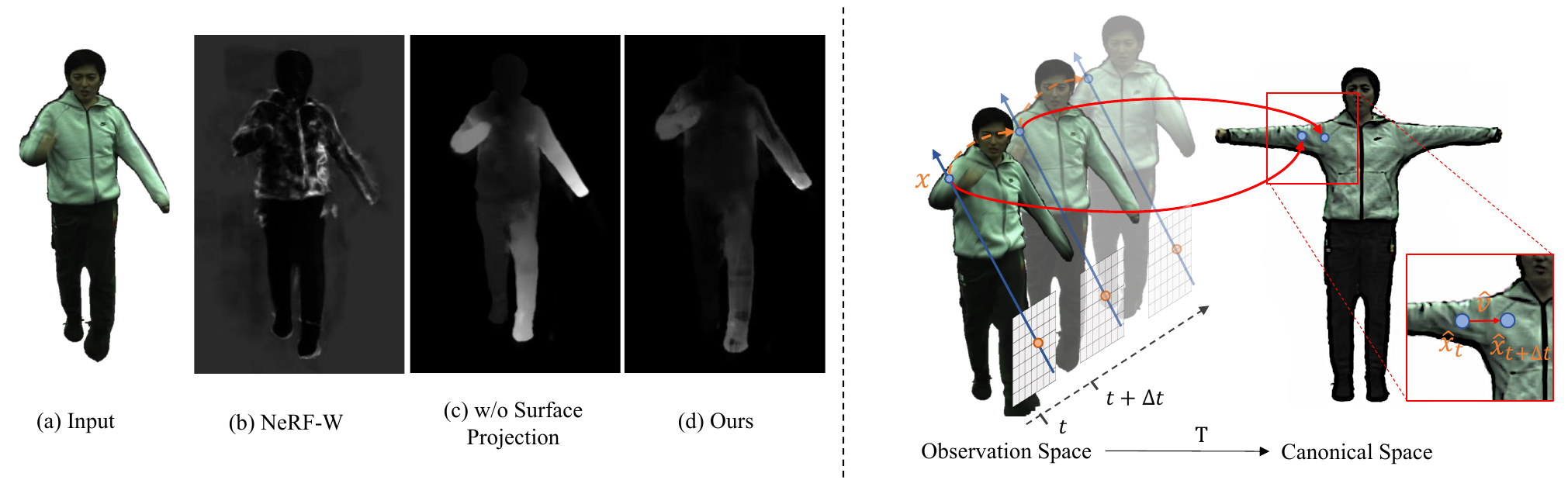}
    \vspace{-15pt}
    \caption{\textbf{Left}: Comparison of different implementation of velocity field. (b) NeRF-W fails to yield meaningful velocity with implicit uncertainty. (c) The velocity without surface projection highlights the wrong regions with less motion blur. \textbf{Right}: Illustration of the computation of the velocity based on the human model in rapid motions.}
    \vspace{-5pt}
    \label{fig:velocity_field}
\end{figure*}

\label{sec:preliminary}
\noindent \textbf{HumanNeRF~\cite{weng2022humannerf}.} 
It is a method derived from NeRF, capable of rendering humans from a monocular video by modeling them as neural fields. In the neural fields, a dynamic human is initially represented within a static canonical space. Then, the human is transformed to observation space by mapping the body into a given pose $\ve{p}$, where color and density are modeled as: 
\begin{equation}
    \ve{c},\sigma = \mathtt{F}(\hat{\ve{x}}),~
    \hat{\ve{x}} = \mathtt{T}(\ve{x}, \ve{p}),
\end{equation}
where $\mathtt{F}$ represents a mapping network that inputs a point $\hat{\ve{x}}$ in the canonical space, returning its color $\ve{c}$ and density $\sigma$. $\mathtt{T}$ denotes the deformation mapping that transposes a point $\ve{x}$ in the observation space to its equivalent coordinates $\hat{\ve{x}}$ within the canonical space. 
Then volumetric rendering is conducted to render the observation ray $\ve{r}$ by aggregating the regressed color and density of the sampled points $\{\ve{x}_i\}_{i=1}^P$ along the ray:
\begin{equation}
\setlength{\abovedisplayskip}{3pt}
\setlength{\belowdisplayskip}{3pt}
\label{eq:human_volume}
\ve{C}(\ve{r}) = \sum_{i=1}^P \mathit{T}_i\alpha_i \ve{c}_i ,
\end{equation}
where $\mathit{T}_i = exp(-\sum_{j=1}^{i-1}\alpha_j\delta_j)$ and $\alpha_i = 1 - \exp{(-\sigma_i\delta_i)}$ denote the transmittance and alpha value of the sampled point $\ve{x}_i$.
$\delta_i$ is the interval between sample $i$ and $i + 1$.


\noindent \textbf{Event Representation.} An event camera is a bio-inspired sensor that asynchronously records changes in intensity. Unlike conventional cameras, which generate frames sequentially at a fixed frame rate, event cameras trigger events independently at each pixel when the intensity change exceeds a constant threshold. This results in properties such as low latency and high dynamic range.
Formally, let $\vm{I}_{u,v}(t)$ represent the instantaneous intensity at pixel coordinate $(u, v)$ at time $t$, and $\vm{L}_{u,v}(t)$ denote its logarithm. An event with polarity $p = \pm1$ is triggered whenever the change in $\vm{L}_{u,v}(t)$ surpasses the threshold $\Theta$, where the polarity indicates the direction of the change (increase or decrease). Following previous works~\cite{rudnev2023eventnerf, qi2023e2nerf}, we apply a single event integral to capture the effective number of events at pixel $(u,v)$ during the time interval $(t_i,t_j]$ as:
\begin{equation}
\label{eq:event_model}
    \hat{E}_{u,v}(t_i,t_j)=\sum_{(t, p):t_i<t \le t_j}p,
\end{equation}
where $t_i$ and $t_j$ are two distinct timestamps sampled within the observed time $[0, t_{max}]$. The value of $\hat{E}_{u,v}(t_i,t_j)$ accounts for the effective number of events at pixel $(u,v)$ during the time interval $(t_i,t_j]$. The threshold $\Theta$ is fixed and symmetric with respect to polarity.

\subsection{Event-oriented Blur-aware Velocity Field}
\label{sec:velocity_field}
The goal of this module is to construct and estimate the velocity field that can reflect the motion blur of humans based on the RGB frames and enhance these regions with event data effectively.
For human reconstruction, consistency across multiple frames is crucial for the precise rendering of appearance and shape. 
However, for the regions of the RGB frames affected by motion blur, the mapping network $\mathtt{F}$ frequently models indistinct color and density, causing the \textit{inconsistency problem} in optimization.
To address this, our approach involves reducing the impact of the blurry regions and refining them with event data to recover the appearance details. 

Previous methods~\cite{martin2021nerf,pan2022activenerf} deploy uncertainty estimation to detect transient regions causing view inconsistency and ambiguity. They represent the radiance values of a scene with a Gaussian distribution, treating the predicted uncertainty as the distribution's variance. Minimizing the Gaussian distribution's negative log-likelihood leads to a scenario where a pixel with high uncertainty is assigned with low influence in the reconstruction. Ideally, pixels with motion blur should be allocated a high variance to lessen their impact on the reconstruction.
However, empirical study reveals that the implicit uncertainty fails to precisely characterize the regions with motion blur (\cref{fig:velocity_field}.b). We contend that the human representation with complex deformation \textit{overfits} the input blurry frames, yielding a lower uncertainty.

To address this issue, we explicitly formulate the uncertainty related to human motion across the multi-frame context to identify the motion blur. Intuitively, motion blur occurs when the pixels observe multiple regions of humans during the exposure interval. Theoretically, based on the human representation in HumanNeRF~\cite{weng2022humannerf}, motion blur occurs when the canonical point $\hat{\ve{x}}$ corresponding to the observation point $\ve{x}$ moves in the exposure interval. Thus, it is reliable to represent the motion blur by employing the velocity of the canonical point $\hat{\ve{x}}$ \textit{w.r.t} timestamp $t$. The large velocity means that the observation point moves fast in the canonical space, which naturally yields motion blur. To achieve this, we formulate a 3D velocity field for the human body in canonical space.
As is shown in \cref{fig:framwork}, by modeling the pose as a function of timestamp $t$, the velocity field $\ve{v}(\ve{x};t)$ can be directly calculated as the derivative of the canonical point $\hat{\ve{x}}$ \textit{w.r.t} timestamp $t$,
\begin{equation}
    \ve{v}(\ve{x};t) = \frac{\partial \hat{\ve{x}}}{\partial t} = \frac{\partial \mathtt{T}(\ve{x}, \ve{p}(t))}{\partial t},
\end{equation}
where $\ve{p}(t)$ is realized using MLP to optimize continuous human poses. In practice, for convenience and fast optimization, we employ finite differences to estimate derivatives: 
\begin{equation}
    \ve{v}(\ve{x};t) \thickapprox (\mathtt{T}(\ve{x}, \ve{p}(t+\Delta t))-\mathtt{T}(\ve{x}, \ve{p}(t)))/\Delta t,
\end{equation}
where $\Delta t$ is a constant time interval.

As shown in \cref{fig:velocity_field} (Right), for the sampled observation point $\ve{x}$, deformation mapping $\mathtt{T}$ deforms it to canonical points of $\hat{\ve{x}}_t$ and $\hat{\ve{x}}_{t+\Delta t}$ corresponding to time $t$ and $t+\Delta t$ respectively.
Also, we find that the motion blur is mostly caused by the motion with velocity horizontal to the human surface, 
we further decompose the initial velocity into normal and tangential components relative to the surface, and apply the tangential component to precisely represent the motion blur. Formally, we project the velocity based on the normal vector $\ve{n}$ of the nearest vertice on the SMPL model~\cite{loper2023smpl} in the canonical space, as is shown in \cref{fig:framwork} (\textit{red rectangle}),
\begin{equation}
\label{eq:projection}
    \hat{\ve{v}}(\ve{x};t) = proj_{\ve{n}}(\ve{v}(\ve{x};t)),
\end{equation}
where $proj_{\ve{n}}(\ve{v}) = \ve{v} - (\ve{v}\cdot \ve{n})\cdot\ve{n}$ projects the vector $\ve{v}$ onto the plane perpendicular to the normal vector $\ve{n}$.
Finally, velocity $V(\ve{r};t)$ in the image space is estimated via volumetric rendering, similar to Eq.~\ref{eq:human_volume},
\begin{equation}
\label{eq:velocity}
\begin{aligned}
    V(\ve{r};t) &= \sum_{i=1}^P T_i\alpha_i \overline{{v}}(\ve{x}_i;t), \\
    \overline{{v}}(\ve{x}_i;t) &=  {v}_0 + \log(1+\exp(||\hat{\ve{v}}(\ve{x}_i;t)||)),
\end{aligned}
\end{equation}
where ${v}_0$ ensures a minimum variance for all the field;  $||*||$ obtains the norm of the velocity vector.
The effectiveness with surface projection of Eq.~\ref{eq:projection} is illustrated in \cref{fig:velocity_field}(d), where the velocity $V(\ve{r};t)$ effectively localizes the region with motion blur.
Without surface projection (\cref{fig:velocity_field}(c)), the velocity highlights the left arm, but it is less blurry than the right hand. This reflects that the velocity on the left hand without surface projection is mostly vertical to the surface.

\noindent \textbf{Discussion.} Prior approaches~\cite{li2024nvfi, wang2023flow} have also accounted for the velocity in deformable scenes. NVFi~\cite{li2024nvfi}, for instance, is geared towards applications such as future frame extrapolation, scene decomposition, and motion transfer. FSDNeRF~\cite{wang2023flow} employs optical flow for supervision, establishing a velocity field within dynamic scenes. 
Differently, our method \textbf{1)} uniquely formulates the velocity in canonical space by considering the modeling of humans, as opposed to the observational space used in prior works~\cite{li2024nvfi, wang2023flow} to derive optical flow, and \textbf{2)} formulates the velocity field for motion blur localization to further enhance with event data.



\subsection{Velocity-Based Rendering Losses}
\label{sec:rendering_loss}

\noindent \textbf{Velocity-Aware Photometric Loss.}
\label{sec:rgb_loss}
Drawing inspiration from NeRF-W~\cite{martin2021nerf}, we also model the color rendering of a ray using a Gaussian distribution $\overline{\ve{C}}(\ve{r})\sim({\ve{C}}(\ve{r}),{\beta}^2(\ve{r}))$, where ${\ve{C}}(\ve{r})$ represents the mean and ${\beta}^2(\ve{r})$ denotes the variance. ${\ve{C}}(\ve{r})$ is determined with rendering result of Eq.~\ref{eq:human_volume}.
The variance ${\beta}^2(\ve{r})$ is explicitly represented with the estimated velocity map $V(\ve{r})$ in Eq.~\ref{eq:velocity}.
Following NeRF-W and ActiveNeRF~\cite{pan2022activenerf}, we enhance the photometric loss by minimizing the negative log-likelihood of the distribution of rays $\ve{r}$ from a batch $\mathbb{R}$,
\begin{equation}
    \boldsymbol{L}_{RGB} = -\sum_{\ve{r}\in \mathbb{R}}\log p(\ve{C}(\ve{r})) = \sum_{\ve{r}\in \mathbb{R}}\frac{||\ve{C}(\ve{r})-\hat{\ve{C}}(\ve{r})||}{2V(\ve{r})} + \frac{\log V(\ve{r})}{2},
\end{equation}
where $t$ is omitted for simplicity, and $\hat{\ve{C}}(\ve{r})$ denotes the ground truth color of camera ray $\ve{r}$.
Given that $V(\ve{r})$ is determined explicitly through Eq.~\ref{eq:velocity} rather than being learned implicitly, it remains constant since we assume the estimated human pose is precise and unchanged.
Thus the $V(\ve{r})$ can be excluded from the objective function. Additionally, following prior arts~\cite{weng2022humannerf, yu2023monohuman} for human reconstruction, we also employ a combination of MSE loss and LPIPS~\cite{zhang2018unreasonable} loss. Consequently, we define our velocity-aware photometric loss as follows:
\begin{equation}
\label{eq:rgb}
    \boldsymbol{L}_{RGB} = \sum_{\ve{r}\in \mathbb{R}}\frac{||\ve{C}(\ve{r})-\hat{\ve{C}}(\ve{r})||}{2V(\ve{r})} + \lambda\sum_{\ve{p}\in \mathbb{P}}\frac{||\mathtt{M}(\ve{C}(\ve{p}))-\mathtt{M}(\hat{\ve{C}}(\ve{p}))||}{2V(\ve{p})},
\end{equation}
where $\ve{p}$ is the sampled patch for the perception model $\mathtt{M}$ of LPIPS, $\lambda$ is the weight balance coefficient.

\noindent \textbf{Velocity-Relative Event Loss.}
\label{sec:event_loss}
While the velocity-aware photometric loss mitigates inconsistencies in regions affected by motion blur, the regions (\eg~, hands) frequently subject to such blur are always assigned high velocity and remain under-constrained. Following previous works~\cite{rudnev2023eventnerf,qi2023e2nerf,ma2023deformable}, the event loss is based on the event integral to supervise the two predicted frames corresponding to the sampled timestamps. 
However, directly applying the event loss can lead to conflicts in areas well-reconstructed with RGB frames, as noise is often present in events. To address this, we introduce a novel velocity-relative event loss to provide adaptive constraints for these regions. The key idea is to apply a relative weight to the event loss based on the velocity value, allowing for targeted supervision of under-constrained regions using the information from well-reconstructed regions.

Given a pixel $(u, v)$, two moments $\{t_i,t_j\}$ are randomly sampled from in the exposure interval of a frame. We take the difference of the predict colors $\{\ve{C}_{t_i},\ve{C}_{t_j}\}$ in the log domain and divide it by the threshold $\Theta$. Then, an estimate of the number of events between two frames for the given pixel is obtained:
\begin{equation}
    {E}_{u,v}(\ve{C}_{t_i},\ve{C}_{t_j}) = \frac{log \ve{C}_{t_i}-log \ve{C}_{t_j}}{\Theta},
\end{equation}
where $\ve{C}_{t_*} = \ve{C}({\ve{r}_{t_*}})$. For simplicity, the ray $\ve{r}$ corresponding to pixel $(u,v)$ is omitted. 
To effectively constrain regions affected by motion blur, we introduce a velocity-relative event loss by \textit{anchoring high-velocity pixels to those with low velocity}, as follows:

\begin{equation}
    \label{eq:event}
    \begin{aligned}
    \boldsymbol{L}_{Event} = \sum_{\ve{r}\in \mathbb{R}} &\beta\mathbb{I}_{\beta > 1} ||{E}_{u,v}(\ve{C}_{t_i},st(\ve{C}_{t_j}))-\hat{E}_{u,v}(t_i,t_j)||^2 + \\
    &\frac{1}{\beta}(1-\mathbb{I}_{\beta > 1}) ||{E}_{u,v}(st(\ve{C}_{t_i}),\ve{C}_{t_j})-\hat{E}_{u,v}(t_i,t_j)||^2
    \end{aligned}
\end{equation}
where $\beta=V_{t_i}/V_{t_j}$ and $\mathbb{I}_{\beta > 1}$ indicates if the  $\beta$ is $>$ 1, and $st(*)$ stop the gradient of the regions with low velocity for optimization. $\hat{E}$ denotes the ground true event integral of Eq~\ref{eq:event_model}.


\subsection{Optimization}
\label{sec:optimization}

\noindent \textbf{Pose Regularization.}
To formulate the velocity as a function of timestamp, the time-to-pose network is jointly trained with the human. To incorporate the human prior to constrain the pose network, we add the pose regularization,
\begin{equation}
\setlength{\abovedisplayskip}{3pt}
\setlength{\belowdisplayskip}{3pt}
    \boldsymbol{L}_{Pose} = \frac{1}{N} \sum_{i=1}^N||\ve{p}(t_i)-\hat{\ve{p}}(t_i)||
\end{equation}
where $\hat{\ve{p}}(t)$ denotes the ground true pose from the frame of the exposure timestamp $t$, which is obtained by manual annotations or a pre-trained estimator; $N$ is the number of the frames.

\noindent \textbf{Velocity-based Alpha Loss.} Following NeuMan~\cite{jiang2022neuman}, we apply regularization to ensure the accumulated alpha map from the human model aligns with the detected human mask. However, since the masks of blurry frames provided by the matting model~\cite{lin2022robust} are also blurred, directly optimizing with them may introduce boundary ambiguity. 
To address this, we extend the alpha loss by incorporating the velocity of the human body, allowing for adaptive optimization of a clearer human boundary:
\begin{equation}
\label{eq:alpha}
    \boldsymbol{L}_{Alpha} = \sum_{\ve{r}\in \mathbb{R}}\frac{||{A}(\ve{r})-\hat{A}(\ve{r})||}{2V(\ve{r})},
\end{equation}
where $A$ represents the predicted mask obtained through volumetric rendering of $\alpha_i$ in the neural field, indicating opacity with a value of 1 and transparency with a value of 0. $\hat{A}$ denotes the estimated alpha mask from pre-trained models. Based on the velocity-based alpha loss, the human mask is optimized with a focus on regions with clear alpha edges, effectively filtering out the blurry alpha values.

\noindent \textbf{Overall Objective.} Incorporating the photometric loss of Eq.~\ref{eq:rgb} and event loss of Eq.~\ref{eq:event}, the overall objective is formulated to optimize the human model, 
\begin{equation}
    \boldsymbol{L} = \boldsymbol{L}_{RGB} + \alpha_e \boldsymbol{L}_{Event} + \alpha_p \boldsymbol{L}_{Pose} + \alpha_a\boldsymbol{L}_{Alpha},
\end{equation}
where $\alpha_e$, $\alpha_p$ and $\alpha_a$ are the are balancing weights.

\section{Experiments}

\begin{table*}[t]

    \centering
    \renewcommand\arraystretch{1.3}
    \caption{Quantitative comparison between our ExFMan and related methods on the ZJU-MoCap~\cite{peng2021neural} dataset. We color cells that have the \colorbox{mycolor1}{best} metric values and \colorbox{mycolor2}{second} ones. Note that LPIPS*=LPIPS $\times 10^3$.}
    \resizebox{1\textwidth}{!}{
    \begin{tabular}{|c|l||ccc||ccc||ccc|}
            \hline
          \multicolumn{2}{|c||}{\multirow{2}{*}{Method}} &\multicolumn{3}{c||}{Subject \textbf{377}}  &  \multicolumn{3}{c||}{Subject \textbf{386}} & \multicolumn{3}{c|}{Subject \textbf{387}}\\
         \cline{3-11}
         \multicolumn{2}{|c||}{} & PSNR$\uparrow$ & SSIM$\uparrow$ & LPIPS*$\downarrow$& PSNR$\uparrow$ & SSIM$\uparrow$ & LPIPS*$\downarrow$& PSNR$\uparrow$ & SSIM$\uparrow$ & LPIPS*$\downarrow$ \\
         \hline
         \multirow{3}{*}{Baselines}& HumanNeRF~\cite{weng2022humannerf} & 18.66 & 0.9463 & 58.50 &22.92&0.9650&42.36&15.45&0.9111&113.87 \\
         \cline{2-11}
         &MonoHuman~\cite{yu2023monohuman} &19.45&0.9500&57.23&22.63&0.9631&49.76&19.56&0.9409&63.34 \\
         \cline{2-11}
         & GauHuman~\cite{hu2024gauhuman} &19.89&0.9467&49.33&23.13&0.9590&\cellcolor{mycolor1}40.09&19.98&0.9332&59.96\\
         \hline
         \multirow{3}{*}{\makecell[c]{RGB-based \\ Deblur}}&MPR~\cite{zamir2021multi}-HumanNeRF~\cite{weng2022humannerf} & 19.05 & 0.9479 & 54.92 &23.08&0.9657&\cellcolor{mycolor2}40.68&19.74&0.9414&61.96 \\
         \cline{2-11}
         &MPR~\cite{zamir2021multi}-MonoHuman~\cite{yu2023monohuman}&19.72&\cellcolor{mycolor2}0.9510&54.86&\cellcolor{mycolor2}23.16&0.9649&45.78&20.00&\cellcolor{mycolor2}0.9428&\cellcolor{mycolor2}59.78\\ 
         \cline{2-11}
         &MPR~\cite{zamir2021multi}-GauHuman~\cite{hu2024gauhuman} &\cellcolor{mycolor2}20.42&0.9487&\cellcolor{mycolor2}46.70&23.06&0.9585&41.01&\cellcolor{mycolor2}20.30&0.9349&\cellcolor{mycolor1}58.84\\
         \hline
        \multirow{4}{*}{\makecell[c]{RGB+Event \\ Deblur}}
        &D2Net~\cite{shang2021bringing}-HumanNeRF~\cite{weng2022humannerf} &18.47&0.9452&59.51&22.31&0.9623&45.70&17.32&0.9279&78.73\\
         \cline{2-11}
         &D2Net~\cite{shang2021bringing}-MonoHuman~\cite{yu2023monohuman} &19.28&0.9496&54.12&23.22&\cellcolor{mycolor2}0.9660&43.32&14.55&0.9223&83.29\\
         \cline{2-11}
         &EFNet~\cite{sun2022event}-HumanNeRF~\cite{weng2022humannerf} & 18.41 & 0.9445 & 55.88 &20.66&0.9545&63.80&15.65&0.9088&112.61\\
         \cline{2-11}
         &EFNet~\cite{sun2022event}-MonoHuman~\cite{yu2023monohuman} &18.77&0.9460&57.36&22.74&0.9631&47.97&19.44&0.9395&62.19\\
         \cline{1-11}
         
         & \textbf{Ours (ExFMan)}&\cellcolor{mycolor1}23.80&\cellcolor{mycolor1}0.9676&\cellcolor{mycolor1}38.61&\cellcolor{mycolor1}24.72&\cellcolor{mycolor1}0.9684&50.46&\cellcolor{mycolor1}22.59&\cellcolor{mycolor1}0.9493&65.08\\
         \hline
        \hline
        \multicolumn{2}{|c||}{\multirow{2}{*}{Method}} &\multicolumn{3}{c||}{Subject \textbf{392}}  &  \multicolumn{3}{c||}{Subject \textbf{393}} & \multicolumn{3}{c|}{Subject \textbf{394}}\\
         \cline{3-11}
         \multicolumn{2}{|c||}{} & PSNR$\uparrow$ & SSIM$\uparrow$ & LPIPS*$\downarrow$& PSNR$\uparrow$ & SSIM$\uparrow$ & LPIPS*$\downarrow$& PSNR$\uparrow$ & SSIM$\uparrow$ & LPIPS*$\downarrow$ \\
         \hline
         \multirow{3}{*}{Baselines}& HumanNeRF~\cite{weng2022humannerf} &16.51&0.9297&81.06&16.90&0.9250&82.41&17.36&0.9326&72.54 \\
         \cline{2-11}
         &MonoHuman~\cite{yu2023monohuman} &16.79&0.9288&87.70&18.26&0.9336&76.96&19.01&0.9403&68.31 \\
         \cline{2-11}
         & GauHuman~\cite{hu2024gauhuman}&17.45&0.9279&71.51&18.06&0.9239&72.03& 18.78&0.9314&61.80\\
          \hline
         \multirow{3}{*}{\makecell[c]{RGB-based \\ Deblur}}
         &MPR~\cite{zamir2021multi}-HumanNeRF~\cite{weng2022humannerf}&16.85&0.9318&78.07& 18.11&0.9329&72.97&18.48&0.9387& 65.92 \\
         \cline{2-11}
         &MPR~\cite{zamir2021multi}-MonoHuman~\cite{yu2023monohuman} &\cellcolor{mycolor2}18.44&\cellcolor{mycolor2}0.9398&73.21&\cellcolor{mycolor2}19.04&\cellcolor{mycolor2}0.9379&68.81&\cellcolor{mycolor2}19.43&\cellcolor{mycolor2}0.9430&63.79\\
         \cline{2-11}
         &MPR~\cite{zamir2021multi}-GauHuman~\cite{hu2024gauhuman} &18.12&0.9327&\cellcolor{mycolor2}68.50&18.91&0.9285&\cellcolor{mycolor2}68.03&19.74&0.9367&\cellcolor{mycolor2}57.71\\
         \hline
        \multirow{4}{*}{\makecell[c]{RGB+Event \\ Deblur}}
        &D2Net~\cite{shang2021bringing}-HumanNeRF~\cite{weng2022humannerf} &16.16&0.9263&85.79&16.76&0.9244&84.33&17.30&0.9308&75.55\\
         \cline{2-11}
         &D2Net~\cite{shang2021bringing}-MonoHuman~\cite{yu2023monohuman} &16.25&0.9322&94.42&14.83&0.9272&86.43&18.59&0.9399&65.01\\
         \cline{2-11}
         &EFNet~\cite{sun2022event}-HumanNeRF~\cite{weng2022humannerf} &16.28&0.9267&80.32&16.99&0.9263&76.40&17.62&0.9337&67.37 \\
         \cline{2-11}
         &EFNet~\cite{sun2022event}-MonoHuman~\cite{yu2023monohuman}&17.51&0.9339&75.39&17.74&0.9302&73.48&18.55&0.9378&66.53 \\
         \cline{1-11}
         
        & \textbf{Ours (ExFMan)}&\cellcolor{mycolor1}22.98&\cellcolor{mycolor1}0.9600&\cellcolor{mycolor1}59.72&\cellcolor{mycolor1} 22.20&\cellcolor{mycolor1}0.9498&\cellcolor{mycolor1}66.98&\cellcolor{mycolor1}22.62&\cellcolor{mycolor1}0.9550&\cellcolor{mycolor1}52.73\\
        
        \hline
    \end{tabular}
    }
\vspace{-6pt}

    \label{tab:zju_mocap_main}
\end{table*}

\subsection{Datasets.}

\noindent \textbf{Synthesis Data.}
We extend the ZJU-MoCap dataset~\cite{peng2021neural} to include simulated motion blur and event data, featuring humans engaged in diverse motions. Following the methodology of HumanNeRF~\cite{weng2022humannerf}, our evaluation primarily focuses on six subjects (377, 386, 387, 392, 393, 394). For training, we use the view from camera 1, while the remaining 22 cameras are reserved for validation. To simulate motion blur and generate sparse event data, we increase the frame rate by interpolating seven additional images between each pair of consecutive frames using the RIFE algorithm~\cite{huang2022rife}. We then produce both blurry frames and event data from these high frame-rate sequences. The blurry frames are created by averaging a set number of sharp images, and the events are generated using the V2E~\cite{hu2021v2e} pipeline. The human poses and segmentation masks are also obtained by averaging the poses and masks corresponding to the sharp images, which are provided by the dataset. 

\noindent \textbf{Real-World Data.}
To capture real-world data, we use the DAVIS346 color event camera~\cite{taverni2018front}, which records spatial-temporally aligned events and RGB frames simultaneously. The camera offers a resolution of 346×260, with an RGB frame exposure time set to 100 ms. Mounted on a tripod, the event camera remains stationary while the subject performs rapid motions (e.g., running, jumping), resulting in motion blur in the RGB frames. To obtain precise and robust human poses in the rapid motion, we apply off-the-shelf EventHPE~\cite{zou2021eventhpe} to estimate the 3D human pose from event data and further refine the pose based on SMPLify~\cite{bogo2016keep} and OpenPose~\cite{8765346}. The segmentation masks are obtained using off-the-shelf RVM~\cite{lin2022robust}. For further details and data capture setup, please refer to the supplementary materials.

\subsection{Evaluation and Implementation Details}
\noindent \textbf{Compared Methods.}
In our comparative evaluation, we select HumanNeRF~\cite{weng2022humannerf}, MonoHuman~\cite{yu2023monohuman}, and GauHuman~\cite{hu2024gauhuman} as baselines for human reconstruction from blurry images. Since there are no existing methods that directly utilize event cameras for this specific task, we design a two-stage comparison pipeline: first, deblurring the RGB images, followed by human reconstruction. The deblurring pipeline is divided into two categories: RGB-based methods and RGB+Event fusion approaches. For the RGB-based methods, we employ MPR~\cite{zamir2021multi}, a leading RGB-based deblurring method, to enhance the initial deblurring process. 
For the RGB+Event fusion methods, we incorporate D2Net~\cite{shang2021bringing} and EFNet~\cite{sun2022event}, two event-guided image deblurring methods, to leverage the potential of integrating event data for image enhancement. After enhancing the frames, we apply HumanNeRF, MonoHuman, and GauHuman for human reconstruction.




\noindent \textbf{Implementation Details.}
We train each scene with 200k iterations on a single NVIDIA A800 GPU. For all data, we set $\alpha_e=0.2$, $\alpha_p=0.01$ and  $\alpha_a=0.01$. We take the batch size as $12$ patches with patch size of $20\times20$ for optimization. 
For event optimization, following E2NeRF~\cite{qi2023e2nerf}, we set the constant threshold $\Theta$ to $0.2$.

\begin{figure*}[t]
    \centering
    \includegraphics[width=1\linewidth]{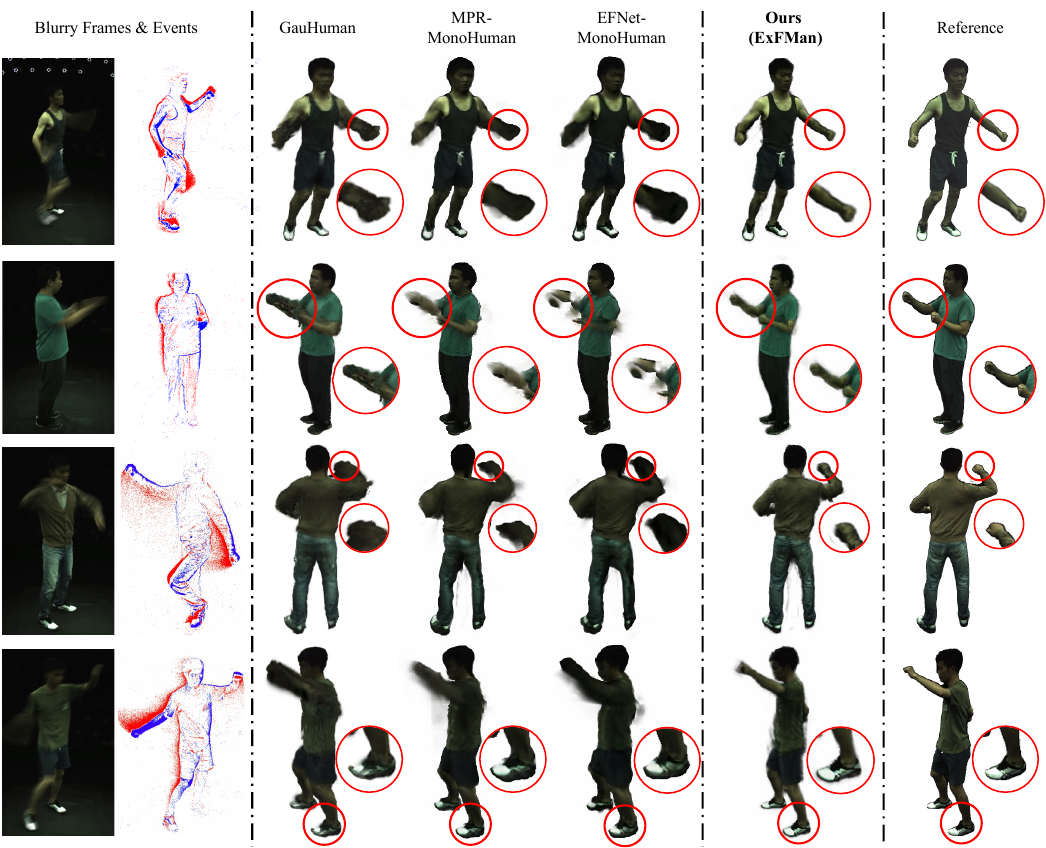}
    \vspace{-20pt}
    \caption{Qualitative results of novel view synthesis in the ZJU-MoCap dataset~\cite{peng2021neural}.}
    \vspace{-20pt}
    \label{fig:zju_mocap_main}
    
\end{figure*}

\subsection{Results on Synthesis Dataset}

For quantitative evaluation, we use metrics such as PSNR, SSIM, and LPIPS~\cite{zhang2018unreasonable} for comparison. The results, presented in \cref{tab:zju_mocap_main}, demonstrate that ExFMan significantly outperforms the baseline methods across various subjects and metrics, particularly in PSNR and SSIM. Although deblurring methods (\eg, MPR-MonoHuman) show some improvements in certain metrics compared to baseline methods (\eg, MonoHuman), their overall performance is limited by their inability to maintain human consistency during the deblurring process. This is likely due to restricted generalization and the absence of effective human modeling for the deblurring methods.
For subject 386, ExFMan achieves the highest performance in PSNR and SSIM, while other comparison methods slightly outperform it in LPIPS. This discrepancy is likely due to the relatively slow motion of the subject, leading to less blur, which the deblurring methods can address to some extent. Additionally, the GauHuman baseline achieves better overall performance than the other two baselines (\ie, HumanNeRF and MonoHuman). Among the two-stage deblurring approaches, MPR-MonoHuman performs better in PSNR and SSIM, while MPR-GauHuman excels in LPIPS. However, both the NeRF-based and 3DGS-based methods exhibit similar degradation effects, indicating that the challenges arise from the data itself, regardless of the representation type used.


A qualitative comparison on the ZJU-MoCap dataset with simulated motion blur highlights the distinctions between ExFMan and other methods, as shown in \cref{fig:zju_mocap_main}. Notably, the GauHuman baseline produces vague boundaries around the human subject and loses detail in motion-heavy regions, such as the hands in the subject 392 ($3^{rd}$ row). The two-stage method MPR-MonoHuman exhibits only modest deblurring, reflecting the limited generalization capabilities of MPR. EFNet-MonoHuman also struggles to capture fine details due to the lack of consistency in the deblurring process, likely because deblurring methods overlook the temporal coherence of the human subject across video frames, as shown in the subject 386 ($2^{nd}$ row).
In contrast, our ExFMan framework optimizes the human model in an end-to-end manner, ensuring global consistency while enhancing detail recovery, particularly in regions affected by motion blur. Additional quantitative and qualitative results are provided in the supplemental material and video.

\subsection{Results on Real-World Dataset}
\begin{figure*}[t]
    \centering
    
    \includegraphics[width=0.96\linewidth]{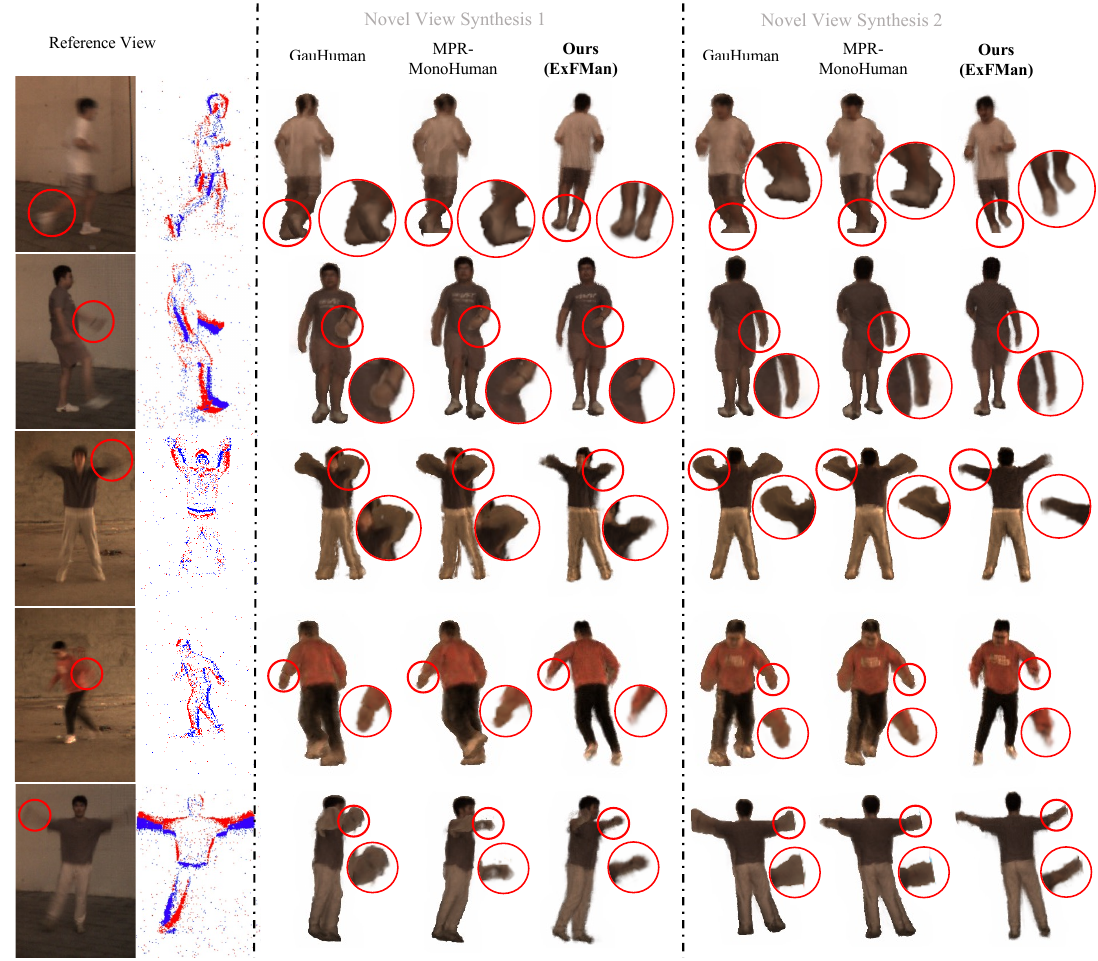}
    \vspace{-20pt}
    \caption{Qualitative results of novel view synthesis in our real-world dataset.}
    \vspace{-10pt}
    \label{fig:real_world}
\end{figure*}

Since the real-world data is captured in-the-wild, involving free and rapid motion to induce motion blur, it is difficult to obtain multi-view ground-truth sharp images for quantitative evaluation. Therefore, we primarily rely on qualitative comparisons, as shown in \cref{fig:real_world}. 
The baseline method (\ie, GauHuman) always produces hazy boundaries and a blurry appearance due to the inherently blurry input frames and masks. For the two-stage method (\ie, MPR-MonoHuman), although minor improvements are shown in certain regions ($5^{th}$ row), it still fails to recover clear human boundaries, largely due to the limited deblurring capacity and generalization of MPR for real-world data. Notably, in the $2^{nd}$ view, the deblurring method directly removes the right hand due to the extreme blur effects.
In contrast, our ExFMan method preserves the color information of the scene and successfully reconstructs sharp human boundaries across all the subjects. For instance, in the $1^{st}$ row, our method clearly delineates the two legs, whereas the comparison methods merge them into a single indistinct part. 
However, the improvement in texture details with our method is less pronounced compared to the ZJU-Mocap dataset, as shown in \cref{fig:zju_mocap_main}. This is due to the limited resolution of the DAVIS346 camera, which makes capturing subtle texture details difficult.
Additional quantitative and qualitative results are provided in the supplemental material and video.
Without ground-truth images, we also provide a no-reference quantitative evaluation to further demonstrate the superiority of our method.

\subsection{Ablation Study and Analysis}
In this section, we conduct ablation experiments by removing each of the components from the ExFMan framework to prove their effectiveness. Additional analysis experiments are provided to demonstrate the robustness of our method.

\begin{table}[t]
    \centering
    \caption{Ablation study on the event-oriented blur-aware velocity field. The results are conducted on subject 377 of ZJU-MoCap~\cite{peng2021neural}. ``VA-PL'' indicates velocity-aware photometric loss. ``VR-EL'' indicates velocity-relative event loss.}
    \renewcommand\arraystretch{1.1}
    \resizebox{ \linewidth}{!}{
    \begin{tabular}{|c||cc||ccc|}
        \hline
        Method & VA-PL & VR-EL & PSNR$\uparrow$ & SSIM$\uparrow$ & LPIPS*$\downarrow$ \\
        \hline
        \hline
        HumanNeRF~\cite{weng2022humannerf} & - & - & 18.66 & 0.9463 & 58.50\\
        \hline
        Baseline  & \ding{55} &  \ding{55}&   20.68 & 0.9568 & 43.55 \\
        \hline
        Baseline w/ VA-PL &   \ding{51} &\ding{55}  &  22.94  & 0.9665&41.42\\
        \hline
        Baseline w/ VR-EL & \ding{55}  & \ding{51} &    21.52 & 0.9615 & 43.33 \\
        \hline
        ExFMan (full) &  \ding{51} & \ding{51} &   
        \textbf{23.80} &\textbf{0.9676} &\textbf{38.61}\\
        \hline
    \end{tabular}
    }
    \label{tab:ablation_loss}
\end{table}

\begin{figure}[t]
    \centering
    
    \includegraphics[width=1\linewidth]{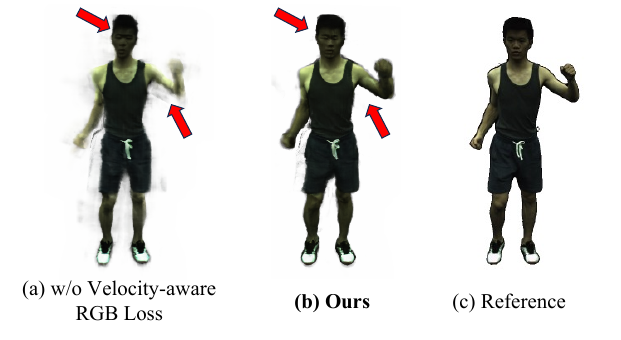}
    \caption{Our velocity-aware photometric loss improves rendering quality at the region with motion blur ((a) vs. (b)). }
    \label{fig:ablation_velocity}
\end{figure}

\begin{figure}[t]
    \centering
    
    \includegraphics[width=1\linewidth]{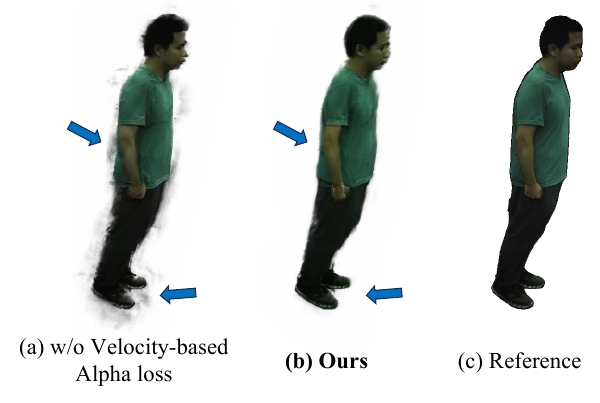}

    \caption{The velocity-based alpha loss obtain clearer boundary compared with baseline ((a) vs. (b)).}
    \label{fig:exfman_blur_loss_ablation}
\end{figure}

\noindent \textbf{Impact of the velocity field.} To reveal the effectiveness of the velocity field. We conduct an ablation study in \cref{tab:ablation_loss}. For the \textit{baseline} method, without a velocity field, we optimize the vanilla photometric loss and event loss, alongside other regularization loss (\ie~pose regularization and velocity-based alpha loss). It is shown that velocity-aware photometric loss obtained significant improvement compared with the baseline. We also show the rendered result of the comparison on photometric loss in \cref{fig:ablation_velocity}, where blur and haze are exhibited on the regions of arms for the baseline method. It is believed that velocity helps to reduce the ambiguity caused by motion blur in the photometric loss and improves the performance. Moreover, velocity-relative event loss also contributes to the performance with supplemental data constrain for the blurry regions, while the full model achieves the best performance.

\noindent \textbf{Impact of the velocity-based alpha loss.} The velocity-based alpha loss is designed to handle motion blur by encouraging the rendered alpha to focus on the clear human mask, taking into account human motion. As shown in \cref{fig:exfman_blur_loss_ablation}, our method achieves clearer boundaries compared to the vanilla alpha loss, while the baseline method tends to produce artifacts around the human subject. These artifacts are a result of optimizing for the blurry human masks.

\noindent \textbf{Impact of event data.} To assess the contribution of event data for our novel task overall the framework, we performed experiments using a baseline model that excludes event data. Consequently, this variant focuses solely on optimizing the velocity-aware photometric and regularization terms (\ie, pose regularization and alpha loss) for the dynamic human. This baseline yields a PSNR of 23.06 dB and an LPIPS of 48.70 on subject 377. Incorporating event data into the model led to significant improvement of \textbf{0.74 dB} in PSNR and \textbf{10.09} in the LPIPS. These results underscore the value of event data in enhancing the reconstruction of humans amidst motion blur.

\begin{figure}[t]
    \centering
    
    \includegraphics[width=1\linewidth]{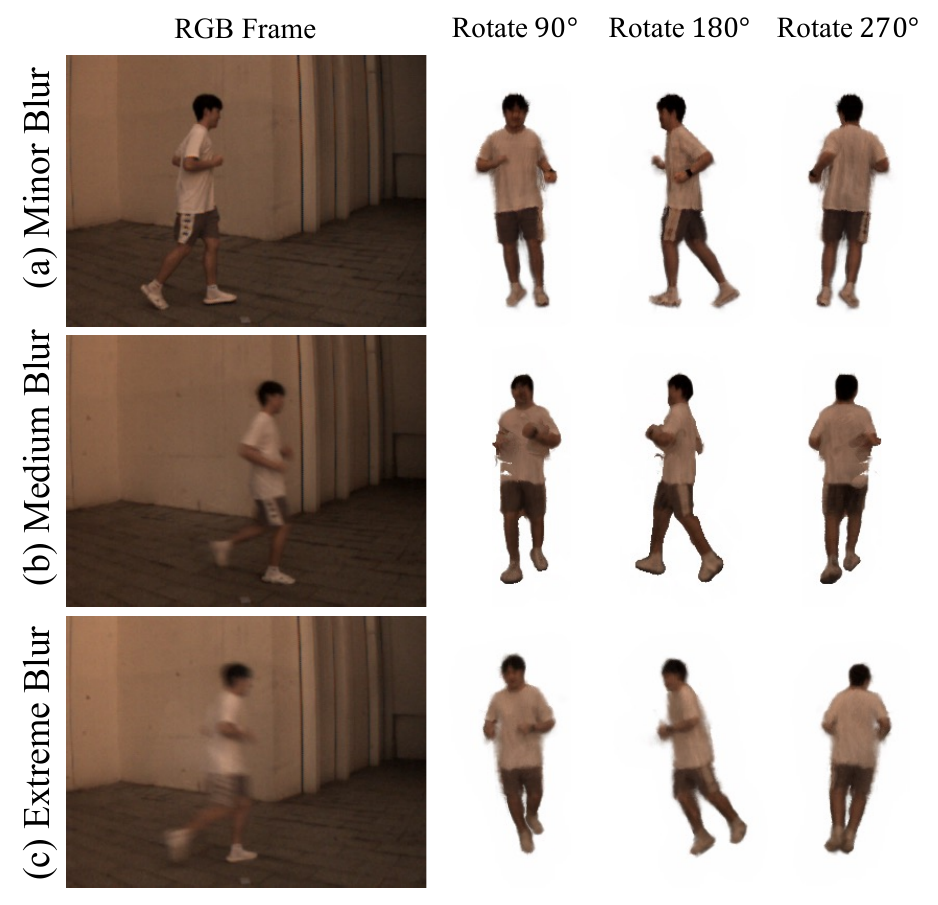}

    \caption{Impact of Motion Blur Intensity. We set the exposure times to 20 ms, 40 ms, and 60 ms, corresponding to minor, medium, and extreme levels of motion blur. Our method demonstrates robustness across varying degrees of motion blur intensity.}
    \label{fig:ablation_blur_degree}
\end{figure}

\noindent \textbf{Impact of motion blur intensity.}
%
In \cref{fig:ablation_blur_degree}, we analyze how the performance of the proposed approach varies with changes in motion blur intensity. By slowing down the human motion and reducing the exposure time, we decrease the blur intensity. Notably, while our method demonstrates superior performance under extreme blur compared to baseline methods (~\cref{fig:real_world}), it also performs effectively in scenarios with less blur. We attribute this robustness and flexibility to the velocity-based loss functions (Eq.~\ref{eq:rgb} and Eq.~\ref{eq:event}), which naturally prioritize clear RGB frames when motion is slower and the frames are sharper. 

\begin{figure}[!]
    \centering
    
    \includegraphics[width=1\linewidth]{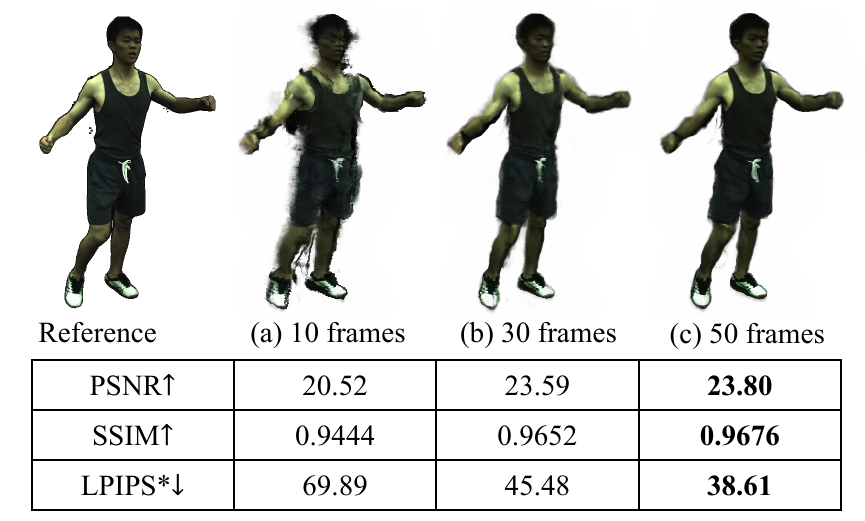}

    \caption{Impact of Sequence Length on Blurry Frames and Event Data. Our method demonstrates robustness when the number of frames exceeds 30.}
    \vspace{-10pt}
    \label{fig:ablation_pose_length}
\end{figure}

\noindent \textbf{Impact of sequence length.}  
In \cref{fig:ablation_pose_length}, we compare the performance of our method across sequences of varying lengths. Starting with the original set of 50 frames and corresponding events, we sample subsets of 10 and 30 frames, along with their associated events between the frames. Notably, the performance only decreased by 0.21 dB in PSNR when reducing from 50 to 30 frames, demonstrating the robustness of our approach to sparse supervision and highlighting the advantage of using events as additional supervision. However, when the number of frames is reduced to 10, the method struggles to reconstruct clear human boundaries, leading to a significant drop in PSNR, as the limited views are insufficient to optimize the entire human structure from the blurry frames.

\section{Discussion and Conclusion}
\noindent \textbf{Discussion.} 
While our method demonstrates promising results in human reconstruction under motion blur, several limitations persist. First, estimating the velocity field inevitably increases computational and memory costs. This issue could be mitigated by employing more advanced human representation techniques, such as hash grids~\cite{muller2022instant} or 3DGS~\cite{kerbl20233d}. 
Second, ExFMan may produce subtle artifacts at the boundaries, primarily due to our reliance on poses and masks derived from pre-trained estimators, which can be affected by rapid motion. Future work could explore developing a more robust human pose and mask estimation method to improve priors in extreme scenarios, using both event and RGB data. Additionally, investigating the reconstruction of dynamic humans based solely on event data is a promising avenue, with the potential to further reduce computational and memory overhead.

\noindent \textbf{Conclusion.}
We propose ExFMan to investigate the issue of rendering humans from motion blur, while previous methods are not feasible due to the inconsistency of human shape and appearance. We incorporate hybrid data of blurry RGB frames and event data, and design a framework based on a novel velocity field to facilitate rendering in motion blur. We demonstrate the effectiveness of the proposed framework on both synthetic datasets and real-world datasets. Our framework shows high-quality appearance recovery and sharp shape reconstruction for human subjects compared with the baselines and state-of-the-art methods.








\clearpage
\bibliographystyle{abbrv-doi-hyperref-narrow}

\bibliography{main}
\end{document}